\documentclass[11pt]{article}
\usepackage[preprint]{acl}

\usepackage{times}
\usepackage{latexsym}
\usepackage[T1]{fontenc}
\usepackage[utf8]{inputenc}
\usepackage{microtype}
\usepackage{graphicx}
\usepackage{enumitem}
\usepackage{booktabs}
\usepackage{amsmath,amssymb}
\usepackage{float}  

\title{LLMs Know the Constraint But Do Not Use It: \\
       Activation Bottlenecks in Pragmatic Constraint Reasoning}

\author{Yubo Li, Ramayya Krishnan, Rema Padman\\
Carnegie Mellon University\\
\{yubol, rk2x, rpadman\}@andrew.cmu.edu}
\begin{document}
\maketitle

\begin{abstract}
When a salient surface cue competes with an implicit feasibility
constraint, LLMs often fail---but aggregate accuracy conflates genuine
constraint inference with conservative defaulting. We formalize the
distinction as \emph{conditional constraint activation}: the constraint
is internally encoded (\textbf{K}nowledge) symmetrically across
constraint-present and -absent prompts (\textbf{S}ymmetry), yet only
sometimes routed into the decision (\textbf{R}outing) and repairable by
a donor activation (Re\textbf{p}air). A quartet diagnostic over 14
models reveals two failure modes; probes on two open weights decode the
constraint above $88\%$, yet activation patching repairs one ($+6.4$
nats) and not the other ($-0.07$). On a mitigation frontier, \emph{no}
prompted intervention reaches the repair corner: all inflate
conservative bias through a single mediation pathway---prerequisite
mention. Hidden-constraint failure is a routing problem, not a
knowledge problem.
\end{abstract}

\section{Introduction}
\label{sec:introduction}

Large language models give plausible-sounding answers to questions
whose implicit prerequisites they ignore. A widely-circulated example
asks whether to walk or drive to a nearby car wash: most frontier
models answer ``walk,'' overlooking that a car wash needs the car to
be present~\citep{knowmadd2026carwash,opper2026carwash,jo2026prompt}.
Behavioral evaluation across heuristic and constraint families
confirms the failure is widespread: under strict aggregation, no model
exceeds 75\% on a 4 heuristic $\times$ 5 constraint
taxonomy~\citep{hob2026}. Yet aggregate accuracy on a constraint-active
item rewards two distinct behaviors equally---a model that infers the
hidden constraint and a model that defaults to a conservative option
for unrelated reasons. The two are dissociable: when the constraint
is \emph{removed}, 12 of 14 evaluated models perform \emph{worse} than
on the original items, with drops up to 39 points~\citep{hob2026}.

\paragraph{This paper.} We argue hidden-constraint failures are best
understood as a \emph{routing} problem rather than a \emph{knowledge}
problem, and provide the diagnostic and mechanistic tools to test the
claim. We formalize \emph{conditional activation} as four falsifiable
conditions on a model $M$ and constraint $C$:
\begin{enumerate}[leftmargin=*,itemsep=-2pt,topsep=0pt]
  \item[\textbf{K.}] \textbf{Knowledge:} a linear probe decodes
    ``constraint applies'' from $M$'s hidden state above $\theta_K$.
  \item[\textbf{S.}] \textbf{Symmetry:} probe accuracy is
    indistinguishable across constraint-active and -removed prompts.
  \item[\textbf{R.}] \textbf{Routing:} probe-projected magnitude
    predicts the gold--shortcut decision logit gap above $\theta_R$.
  \item[\textbf{P.}] \textbf{Repair:} patching hidden states from an
    explicit-constraint prompt restores correctness without flipping
    constraint-removed pairs.
\end{enumerate}
A model satisfying K, S, P but failing R exhibits a \emph{conditional
activation bottleneck}: the constraint is internally represented but
not routed into the decision.

\paragraph{Two failure modes, mechanistically dissociated.}
A quartet-style evaluation---an \textsc{Active} item, a paired
\textsc{Removed} control, an \textsc{Explicit} variant, and a
length- and frame-matched \textsc{Salience\,Control}---reveals two
populations across 14 models.
Over-activation models (Llama-4, Claude Opus 4.6, Kimi K2.5,
Qwen3.5-27B, GPT-5.2, Claude Sonnet 4.5, Gemini 3 Pro) give the
constraint-heavy answer even when the constraint is absent (CBI
$+0.13$ to $+0.29$). Under-activation models (GPT-OSS-\{20B, 120B\})
under-apply the constraint when required (CBI $-0.10$ to $-0.19$).
On the two open weights, the modes split mechanistically: in
Qwen3-14B, patching constraint information from an Explicit prompt
into a failing Active prompt shifts the gold--shortcut gap by $+6.4$
nats while leaving Removed pairs essentially unchanged ($-0.84$); in
GPT-OSS-20B, an $88\%$-accurate probe at $L_{20}$ confirms the
constraint is internally available, yet patching moves the answer by
only $-0.07$ nats. The under-activation mode is the textbook routing
failure; the over-activation mode is a prior-bias problem with
routing intact.

\paragraph{Salience-controlled hint activation.}
A common reading of the explicitness-gradient observation~\citep{hob2026}
is that hints ``remind'' the model of the constraint. We test this by
inserting a length- and frame-matched neutral filler in place of the
constraint cue. The Salience-Adjusted Hint Gain ranges from $+0.021$
(Gemini~3~Pro---no specificity above matched salience) to $+0.294$
(Llama-4---strong specificity), with matched controls flat across all
five ladder levels.

\paragraph{The mitigation frontier.} We re-evaluate four prompted
interventions---CoT~\citep{wei2022chain,kojima2022large},
precondition listing, goal decomposition, and counterfactual
checking~\citep{wang2023selfconsistency}---on a two-dimensional
frontier (active-item gain vs.\ removed-pair harm). Across all 40
(model $\times$ strategy) cells, strategies cluster in the high-harm,
near-zero-gain region: mean removed-pair harm $+0.44$ to $+0.47$ in
CBI units, mean active gain only $+0.01$ to $+0.04$; \emph{none}
reaches the repair corner. They converge on a shared mediation
pathway---mediated-correctness share $\geq\!0.91$ in every cell---each
working by inducing the same surface behavior (prerequisite mention)
that boosts Active-correctness while over-triggering on Removed
prompts. A reasoning-budget sweep on four thinking-mode models
reproduces the same harm-without-repair signature~\citep{snell2024scaling,muennighoff2025s1}.

\paragraph{Contributions.}
(i)~A formal definition of \emph{conditional constraint activation}
as a falsifiable K/S/R/P claim, instantiated as a quartet diagnostic
with salience controls. (ii)~Empirical evidence for two distinct
failure modes and a mechanistic dissociation via probes and patching
on two open weights. (iii)~The \emph{mitigation frontier} protocol,
showing that all four established prompted methods move along the
harm axis. (iv)~A measured mediation chain explaining \emph{why}
prompted compute helps when it does. Together: hidden-constraint
failure is a circuit-level conditional-activation problem; prompted
mitigations cannot repair it; the remedy must operate on the routing
direction itself.

\section{Related Work}
\label{sec:related}

\paragraph{Heuristic shortcuts and the knowing-vs-using gap.}
LLMs follow surface-level statistical regularities rather than perform
the intended
computation~\citep{geirhos2020shortcut,mccoy2019right,nikankin2024arithmetic},
showing content effects~\citep{lampinen2024language} and easy
distractibility~\citep{shi2023large,mirzadeh2024gsm}. A parallel line
shows models internally represent information they do not output:
latent multi-hop chains~\citep{yang2024large}, the reversal
curse~\citep{berglund2024reversal}, hallucinations whose truth status
is encoded in hidden state~\citep{orgad2025llms}, and truth-direction
inference-time interventions~\citep{li2024inferencetime,zou2023representation}.
The Heuristic Override Benchmark~\citep{hob2026} extended the
behavioral failure to pragmatic reasoning across four heuristic
families and five constraint families, but leaves open whether a
correct answer reflects genuine constraint inference or conservative
defaulting. We push past that ambiguity with a quartet-based
diagnostic and with mechanistic experiments that localize the failure
to a specific routing pathway.

\paragraph{Linear probes and activation patching.}
Linear classifiers on intermediate
representations~\citep{alain2017understanding,conneau2018what,hewitt2019structural,belinkov2022probing}
are the standard test for whether a feature is present in hidden state,
though high probe accuracy does not by itself establish causal
use~\citep{hewitt2019control}. Many factual associations and
truth-related axes are linearly
decodable~\citep{petroni2019language,geva2021transformer,dai2022knowledge,gurnee2024language,burns2023discovering,zou2023representation,orgad2025llms}.
Causal mediation~\citep{vig2020investigating,geiger2021causal} and
activation
patching~\citep{meng2022rome,wang2023ioi,hanna2023how,lieberum2023does,wu2023interpretability,marks2024sfc,zhang2024best,heimersheim2024how}
have localized circuits for factual recall, indirect-object
identification, and arithmetic~\citep{stolfo2023mechanistic}. Most
prior patching work targets \emph{factual recall} or \emph{single-step
symbolic} computations. We apply patching to \emph{conditional pragmatic
reasoning} on two open weights with opposite behavioral failure modes,
yielding a mechanistic dissociation aggregate behavior obscures.

\paragraph{Prompted mitigations and inference-time compute.}
A steady stream of prompting interventions has been introduced---
chain-of-thought~\citep{wei2022chain,kojima2022large},
self-consistency~\citep{wang2023selfconsistency},
tree-of-thoughts~\citep{yao2023tree},
plan-and-solve~\citep{wang2023plansolve},
self-refine~\citep{madaan2023selfrefine},
verifier reranking~\citep{cobbe2021verifiers,lightman2024letsverify},
reflexion~\citep{shinn2023reflexion},
step-back~\citep{zheng2024takeastep}---alongside long-CoT thinking-mode
models~\citep{openai2024o1,deepseek2025r1,anthropic2025extendedthinking}
and inference-scaling work~\citep{snell2024scaling,brown2024monkeys,muennighoff2025s1}.
Counter-evidence shows LLMs cannot reliably self-correct
reasoning~\citep{huang2024large}, CoT benefits are narrow~\citep{sprague2025cot},
and overthinking can degrade
performance~\citep{chen2025think23}. Prior reports rarely audit
whether constraint-removed minimal pairs are damaged by the same
intervention. Our mitigation frontier re-evaluates four established
strategies on a two-dimensional axis (active gain vs.\ pair harm),
showing that gains in the literature are routinely along the
conservative-bias axis rather than the repair axis.

\paragraph{Conservative bias and answer asymmetries.}
Sycophancy~\citep{perez2023discovering,sharma2024towards,wei2024simple},
hedging~\citep{zhou2024relying,xiong2024can}, and calibration
asymmetries~\citep{kadavath2022language} are documented RLHF-induced
behaviors that shift outputs toward safe options. Our over-activation
failure mode is their mechanistic counterpart---a constraint-heavy
default that is correct when the constraint applies and wrong when it
does not. Diagnosing it behaviorally requires the minimal-pair
contrast; diagnosing it mechanistically requires checking whether the
constraint signal in hidden state actually shifts the decision under
patching.

\section{Methods}
\label{sec:methods}

We instantiate the four conditional-activation conditions
(\textbf{K}/\textbf{S}/\textbf{R}/\textbf{P}, \S\ref{sec:introduction})
through three measurement layers: (i)~a behavioral quartet diagnostic
with matched salience controls, (ii)~a salience-controlled hint ladder,
and (iii)~hidden-state probes plus activation patching on open-weight
models. We additionally re-evaluate four prompted mitigations on a
two-axis frontier and probe inference-time compute through a budget
sweep on four thinking-mode models.

\subsection{Datasets and Conditions}
\label{sec:methods:datasets}

\paragraph{Item universe.} We start from the HOB benchmark
\citep{hob2026}: a 4-heuristic $\times$ 5-constraint taxonomy
($4 \times 5 = 20$ cells) over $\approx$140 scenarios, each a goal +
salient surface cue + hidden constraint $C$ that, when active, makes
the cue-following answer infeasible. We use \textsc{core100}: 100 base
scenarios stratified across all 20 cells.

\paragraph{Quartet construction.}
For each scenario we construct four conditions paired by scenario id
(Table~\ref{tab:quartet_methods}). \textsc{Salience\,Control} controls
for the alternative hypothesis that explicit-constraint gains are a
non-specific salience effect: fillers are length-bucketed, neutral
declarative sentences, assigned hash-deterministically by instance id.

\begin{table}[t]
\small
\centering
\setlength{\tabcolsep}{3pt}
\begin{tabular}{@{}l p{5.0cm}@{}}
\toprule
\textbf{Condition} & \textbf{Definition} \\
\midrule
\textsc{Active} & Original scenario; $C$ satisfied implicitly. \\
\textsc{Removed} & Constraint-removed counterfactual; cue preserved. \\
\textsc{Explicit} & \textsc{Active} $+$ $C$ as one declarative sentence. \\
\textsc{Sal.\,Ctrl} & \textsc{Active} $+$ length-/frame-matched neutral filler at the same position. \\
\bottomrule
\end{tabular}
\caption{The four quartet conditions per scenario.}
\label{tab:quartet_methods}
\end{table}

\paragraph{Hint ladder.} A five-level ladder $L_0\!\prec\!\cdots\!\prec\!L_4$,
where $L_0$ is \textsc{Active} and $L_4$ approaches \textsc{Explicit}.
At every level $L_i$ ($i\!\geq\!1$) we inject three matched negative
controls: \textsc{ctrl\_lex}, \textsc{ctrl\_len}, \textsc{ctrl\_heur}.

\paragraph{Sample sizes.} $T$ trials per (model, condition): $T\!=\!10$ for
API models, $T\!=\!8$ for local-GPU models, $T\!=\!4$ for the budget sweep.

\subsection{Models}
We evaluate 14 LLMs spanning four providers (Claude Opus/Sonnet, GPT
5.2/5.4/OSS-20B/OSS-120B, Gemini~3~Pro, Grok 4.2, DeepSeek-R1, Kimi
K2.5, Llama-4, Qwen3-14B/32B, Qwen3.5-27B). For probes and patching
we use \textbf{Qwen3-14B} (balanced) and \textbf{GPT-OSS-20B}
(under-activation), the two open weights spanning the failure modes.

\subsection{Behavioral Metrics}
\label{sec:methods:metrics}
Let $\mathrm{Acc}_c$ denote accuracy in condition $c$.
\begin{equation*}
\begin{aligned}
\mathrm{CBI} &= \mathrm{Acc}_{\textsc{A}} - \mathrm{Acc}_{\textsc{R}} \\
\mathrm{SalAdjGain} &= \mathrm{Acc}_{\textsc{E}} - \mathrm{Acc}_{\textsc{SC}} \\
\mathrm{PCA} &= \tfrac{1}{|S|}\sum_{s}\widehat{\mathrm{Acc}}_{\textsc{A}}(s)\!\cdot\!
                                       \widehat{\mathrm{Acc}}_{\textsc{R}}(s) \\
\mathrm{CAS} &= \mathrm{PCA}\cdot(1 - |\mathrm{CBI}|)
\end{aligned}
\end{equation*}
$\mathrm{CBI}\!>\!\theta_{\text{CBI}}\!=\!0.10$ flags over-activation;
$\mathrm{CBI}\!<\!-\theta_{\text{CBI}}$ flags under-activation. Over-
and under-activation rates are $\mathrm{OAR}\!=\!1\!-\!\mathrm{Acc}_{\textsc{R}}$
and $\mathrm{UAR}\!=\!1\!-\!\mathrm{Acc}_{\textsc{A}}$. The ladder
\emph{specificity gap} at level $i$ is $\mathrm{Acc}_{L_i} -
\max_{\tau}\mathrm{Acc}_{L_i^{(\tau)}}$; the
minimum-effective-hint-level (MEHL) is the smallest $i$ at which it
exceeds a preregistered $0.10$ margin.

\subsection{Mitigation Frontier}
\label{sec:methods:mitigation}
We evaluate four prompted strategies grouped by hypothesized
mechanism: Class A raises the conservative threshold (vanilla CoT
\citep{wei2022chain,kojima2022large}); Class B induces prerequisite
search (\textit{precondition listing}, \textit{goal decomposition});
Counterfactual-check
\citep{wang2023selfconsistency} is an oracle upper bound.
Each is a single-prompt prefix; every other condition is held fixed.
For (model, strategy) we report the
two-axis frontier
$\textsc{ActiveGain}\!=\!\mathrm{Acc}^{\text{strat}}_{\textsc{A}}
   -\mathrm{Acc}^{\text{zero}}_{\textsc{A}}$ vs.\
$\textsc{PairHarm}\!=\!\mathrm{CBI}^{\text{strat}} -
\mathrm{CBI}^{\text{zero}}$; the repair corner (positive
\textsc{ActiveGain}, non-positive \textsc{PairHarm}) sits at the
bottom-right.

\subsection{Reasoning-Budget Sweep}
We sweep $\text{budget}\in\{256,1024,4096,16384\}$ thinking tokens on
the four thinking-mode models. Provider-specific budget routing
(\texttt{thinking.budget\_tokens}, \texttt{reasoning\_effort},
\texttt{ThinkingConfig}) is detailed in
Appendix~\ref{app:repro}.

\subsection{Mediation Analysis}
\label{sec:methods:mediation}
We estimate the chain
\textit{intervention $\to$ prerequisite mention $\to$ correctness}.
Mention is detected by content-word overlap with the canonical hidden
constraint (Appendix~\ref{app:repro}). We report
$\Pr(\text{c}\mid\text{m})$, $\Pr(\text{c}\mid\neg\text{m})$,
$\Pr(\text{m})$, and the \emph{mediated correctness share}, the
fraction of correct answers attributable to mentioning traces.

\subsection{Hidden-State Probes (K, S, R)}
\label{sec:methods:probes}
For Qwen3-14B and GPT-OSS-20B we cache the post-block residual hidden
state at the final input token across all transformer layers on every
\textsc{core100} quartet item. We train one per-layer logistic
regression (\texttt{liblinear}, $C\!=\!1$, with per-feature
standardization) to predict \textsc{Active} vs.\ \textsc{Removed}.
Cross-validation is \emph{scenario-grouped} 5-fold: the four
conditions for a scenario co-occur in the same fold.
\textbf{K} is satisfied at the best layer when CV accuracy exceeds
$\theta_K\!=\!0.80$. \textbf{S} (symmetry) is the test-accuracy gap
between balanced Active-only and Removed-only training subsets,
threshold $\theta_S\!=\!0.05$. \textbf{R} (routing) is the Spearman
correlation between probe projection and gold--shortcut decision
logit gap at the best probe layer, threshold $\theta_R\!=\!0.30$.

\subsection{Activation Patching (P)}
\label{sec:methods:mechanism}
For each scenario we forward the \textsc{Explicit} (donor) prompt and
cache the final-token hidden state $h^{\text{donor}}_{\ell^\star}$ at
the probe's best layer $\ell^\star$. We then forward the recipient
(\textsc{Active} or \textsc{Removed}) prompt with a forward hook on
$\ell^\star$ that overwrites the final-token output with
$h^{\text{donor}}_{\ell^\star}$, and recompute the gold--shortcut log-
probability gap. The patched-minus-baseline difference $\Delta$
measures how much the constraint-encoding activation changes the
decision. \textbf{P} requires specificity: $\Delta^{\text{Active}}\!>\!
\theta_P\!=\!1.0$ nat \emph{and} $|\Delta^{\text{Removed}}|\!\leq\!\theta_P$,
preregistered against the typical logit-gap magnitude
($\sigma_\Delta\!\approx\!3$ nats). Implementation, donor-shuffling
control, and layer/position sweeps are in Appendix~\ref{app:repro}.

\section{Results}
\label{sec:results}

We present results in the order of the four conditions:
\textbf{K}~(knowledge present),
\textbf{S}~(symmetric across paired conditions),
\textbf{R}~(routed into the decision)---all in \S\ref{sec:results:K}---
and \textbf{P}~(repair via patching, \S\ref{sec:results:P}).
\S\ref{sec:results:behavior} reports the behavioral signature on all
14~models; \S\ref{sec:results:ladder} reports the salience-controlled
hint ladder; \S\ref{sec:results:frontier} reports the mitigation
frontier; \S\ref{sec:results:budget} reports the reasoning-budget sweep
with mediation analysis.

\subsection{Behavioral signature: two failure modes}
\label{sec:results:behavior}

\begin{figure}[t]
\centering
\includegraphics[width=\linewidth]{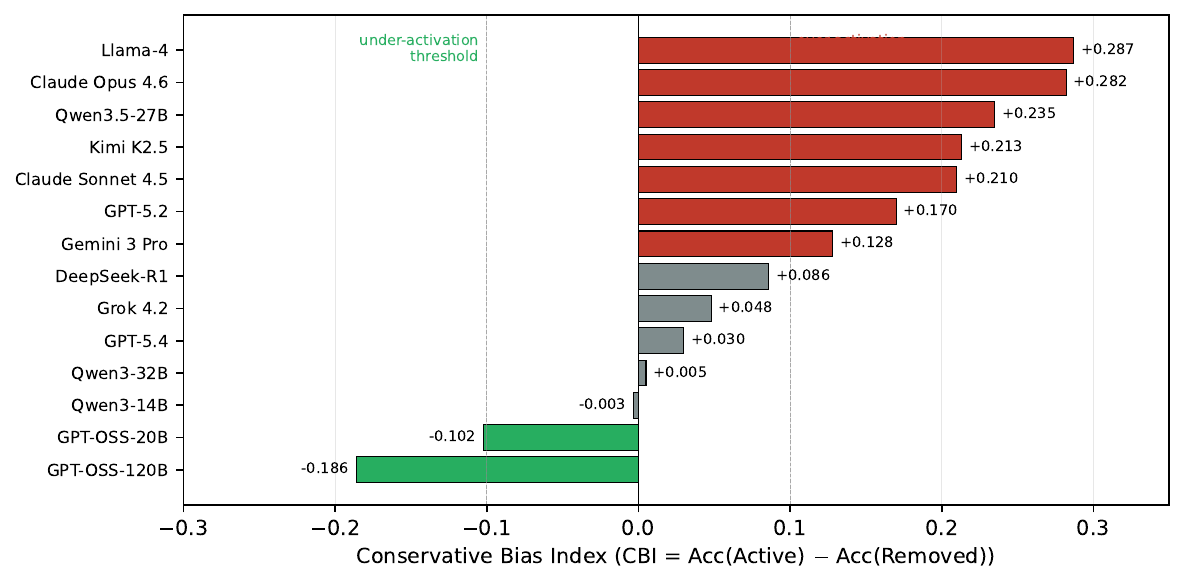}
\caption{Per-model Conservative Bias Index, rank-ordered. Two
populations are separated by $|\text{CBI}|=0.10$ (grey dashed):
over-activation (red, 7 models), balanced (grey, 5 models), and
under-activation (green, 2 models).}
\label{fig:cbi_bars}
\end{figure}

Table~\ref{tab:quartet} reports per-model accuracy in all four
quartet conditions, plus the derived metrics. Across 14~models we
observe \emph{two} populations separated by the sign of CBI
(Fig.~\ref{fig:cbi_bars}).

\paragraph{Over-activation (CBI~$>\!+0.10$): 7 models.}
Llama-4, Claude Opus 4.6, Qwen3.5-27B, Kimi K2.5, Claude Sonnet 4.5,
GPT-5.2, and Gemini 3 Pro (CBI $+0.13$ to $+0.29$; see
Table~\ref{tab:quartet}). These models pick the constraint-heavy
answer even when the constraint is absent: Removed accuracy falls
11--33 points below Active. Pair-Consistent Accuracy correspondingly
collapses (e.g.\ $0.342$ for Llama-4 vs.\ $0.792$ marginal Active
accuracy).

\paragraph{Balanced (|CBI|~$\leq\!0.10$): 5 models.}
DeepSeek-R1 ($+0.086$), Grok 4.2 ($+0.048$), GPT-5.4 ($+0.030$),
Qwen3-32B ($+0.005$), Qwen3-14B ($-0.003$). Active and Removed
accuracies are within $9$ points across all five.

\paragraph{Under-activation (CBI~$<\!-0.10$): 2 models.}
GPT-OSS-20B ($-0.103$) and GPT-OSS-120B ($-0.186$). These models are
\emph{more} accurate on Removed than on Active---they fail to apply
the constraint when it matters.

\begin{table*}[t]
\small
\centering
\setlength{\tabcolsep}{4pt}
\begin{tabular}{lrrrrrrrr}
\toprule
& \multicolumn{4}{c}{Per-condition Acc} & & & & \\
\cmidrule(lr){2-5}
\textbf{Model} & Active & Removed & Explicit & Sal\,Ctrl
              & \textbf{CBI} & PCA & CAS & SalAdjGain \\
\midrule
\multicolumn{9}{l}{\emph{Over-activation}} \\
Llama-4             & 0.792 & 0.505 & 0.926 & 0.632 & $+0.287$ & 0.342 & 0.400 & $+0.294$ \\
Claude Opus 4.6     & 0.886 & 0.604 & 0.923 & 0.833 & $+0.282$ & 0.523 & 0.535 & $+0.090$ \\
Qwen3.5-27B         & 0.925 & 0.690 & 0.969 & 0.851 & $+0.235$ & 0.645 & 0.638 & $+0.118$ \\
Kimi K2.5           & 0.928 & 0.715 & 0.956 & 0.853 & $+0.213$ & 0.650 & 0.664 & $+0.103$ \\
Claude Sonnet 4.5   & 0.837 & 0.627 & 0.936 & 0.766 & $+0.210$ & 0.496 & 0.525 & $+0.170$ \\
GPT-5.2             & 0.824 & 0.654 & 0.929 & 0.802 & $+0.170$ & 0.513 & 0.539 & $+0.127$ \\
Gemini 3 Pro        & 0.907 & 0.779 & 0.910 & 0.889 & $+0.128$ & 0.693 & 0.707 & $+0.021$ \\
\midrule
\multicolumn{9}{l}{\emph{Balanced}} \\
DeepSeek-R1         & 0.850 & 0.764 & 0.957 & 0.795 & $+0.086$ & 0.627 & 0.649 & $+0.162$ \\
Grok 4.2            & 0.883 & 0.835 & 0.954 & 0.790 & $+0.048$ & 0.725 & 0.737 & $+0.164$ \\
GPT-5.4             & 0.822 & 0.792 & 0.938 & 0.723 & $+0.030$ & 0.634 & 0.651 & $+0.215$ \\
Qwen3-32B           & 0.771 & 0.766 & 0.932 & 0.667 & $+0.005$ & 0.580 & 0.591 & $+0.265$ \\
Qwen3-14B           & 0.762 & 0.765 & 0.944 & 0.694 & $-0.003$ & 0.567 & 0.583 & $+0.250$ \\
\midrule
\multicolumn{9}{l}{\emph{Under-activation}} \\
GPT-OSS-20B         & 0.765 & 0.868 & 0.941 & 0.694 & $-0.103$ & 0.657 & 0.664 & $+0.248$ \\
GPT-OSS-120B        & 0.686 & 0.872 & 0.857 & 0.627 & $-0.186$ & 0.595 & 0.598 & $+0.230$ \\
\bottomrule
\end{tabular}
\caption{Quartet behavioral metrics on \textsc{core100}, 14 models, sorted
by CBI. PCA is the soft (per-scenario expectation) variant; CAS combines
PCA with the conservative-bias penalty. SalAdjGain is the
explicit-minus-salience-control accuracy gap.}
\label{tab:quartet}
\end{table*}

\paragraph{Salience-Adjusted Hint Gain.} SalAdjGain ranges from
$+0.021$ (Gemini~3~Pro: no specificity beyond matched salience) to
$+0.294$ (Llama-4) and $+0.265$ (Qwen3-32B). The under-activation
models retain large positive SalAdjGain (GPT-OSS-20B: $+0.248$;
GPT-OSS-120B: $+0.230$), confirming the constraint \emph{can} be
activated when made explicit. Gemini~3~Pro's near-zero SalAdjGain
demonstrates that a model can show large \emph{nominal} hint gains
entirely explained by added textual salience.

\subsection{Hint ladder with matched controls}
\label{sec:results:ladder}

On the six API models with full ladder collection, explicit-condition
accuracy rises monotonically with hint level for every model; the
three matched controls remain within $\pm 0.04$ of $L_0$ across all
four ladder rungs. The specificity gap exceeds the preregistered
$0.10$ margin by $L_2$ for over-activation models and by $L_3$ for
balanced models (DeepSeek-R1, GPT-5.4). Per-model curves and
specificity gaps are in Figs.~\ref{fig:ladder} and
\ref{fig:ladder_specgap} (Appendix).

\subsection{The mitigation frontier}
\label{sec:results:frontier}

Table~\ref{tab:mit_frontier} reports per-strategy means over 10 API
models; full per-(model, strategy) detail is in
Table~\ref{tab:mit_frontier_full} (Appendix).

\paragraph{Existing strategies inflate CBI without repairing routing.}
Across all 40 (model, strategy) cells, \textsc{PairHarm} is large
($+0.28$ to $+0.65$) and \textsc{ActiveGain} is small ($-0.08$ to
$+0.11$); none reaches the repair corner. Counter to our preregistered
hypothesis, Class-B interventions (\textit{precondition\_listing},
\textit{goal\_decomposition}) do not move closer to repair than
Class-A: mean \textsc{PairHarm} $+0.44$ vs.\ $+0.47$, mean
\textsc{ActiveGain} $+0.038$ vs.\ $+0.012$. The
\textit{counterfactual\_check} oracle shows the same pair-harm
signature ($+0.44$ mean). Llama-4 is the worst case: \emph{every}
strategy reduces its Active accuracy ($-0.018$ to $-0.082$).

\begin{figure*}[t]
\centering
\includegraphics[width=0.95\linewidth]{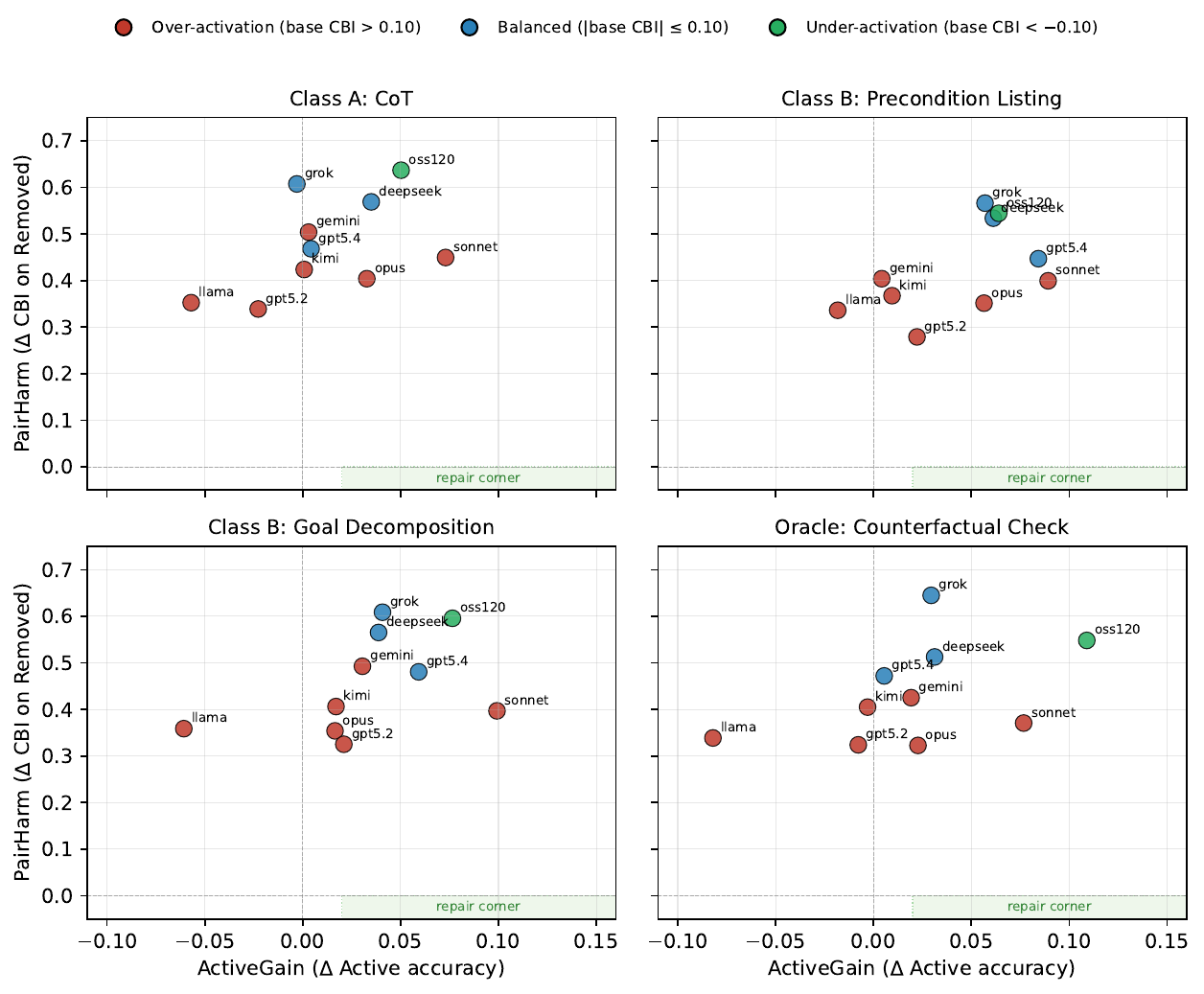}
\caption{Mitigation frontier across ten API models and four prompted
strategies. Each point is one (model, strategy) cell at the model's
own zero-shot baseline. The green ``repair corner'' (bottom-right of
each panel: positive ActiveGain, non-positive PairHarm) is empty for
every strategy. Points colored by failure mode of the base model.}
\label{fig:frontier}
\end{figure*}

\paragraph{Mediation through prerequisite mention.}
Fig.~\ref{fig:frontier} visualizes the per-strategy frontier; every
point sits in the high-harm, low-gain region.

Across all 40 cells, the conditional gap
$\Pr(\text{c}\mid\text{m}) - \Pr(\text{c}\mid\neg\text{m})$ is large
($+0.34$ to $+0.64$), $\Pr(\text{m})$ is high ($0.84$ to $0.96$), and
the mediated-correctness share is $0.91$ to $0.99$. Every strategy
``works'' by inducing the same surface behavior (prerequisite
mention); they differ only in how much they over-trigger it on
Removed items.

\begin{table}[t]
\small
\centering
\setlength{\tabcolsep}{4pt}
\begin{tabular}{@{}l rrr@{}}
\toprule
\textbf{Strategy} & ActGain & PairHarm & MedShare \\
\midrule
cot (A)                 & $+0.012$ & $+0.475$ & 0.962 \\
precond.\ listing (B)   & $+0.043$ & $+0.423$ & 0.974 \\
goal decomp.\ (B)       & $+0.034$ & $+0.458$ & 0.975 \\
counterfactual (oracle) & $+0.020$ & $+0.436$ & 0.954 \\
\bottomrule
\end{tabular}
\caption{Per-strategy means over 10 API models. Every strategy
moves overwhelmingly along the harm axis; ActiveGain is small and
indistinguishable across classes. Per-(model, strategy) detail is in
Table~\ref{tab:mit_frontier_full} (Appendix).}
\label{tab:mit_frontier}
\end{table}

\subsection{Reasoning-budget sweep}
\label{sec:results:budget}

\begin{figure}[t]
\centering
\includegraphics[width=\linewidth]{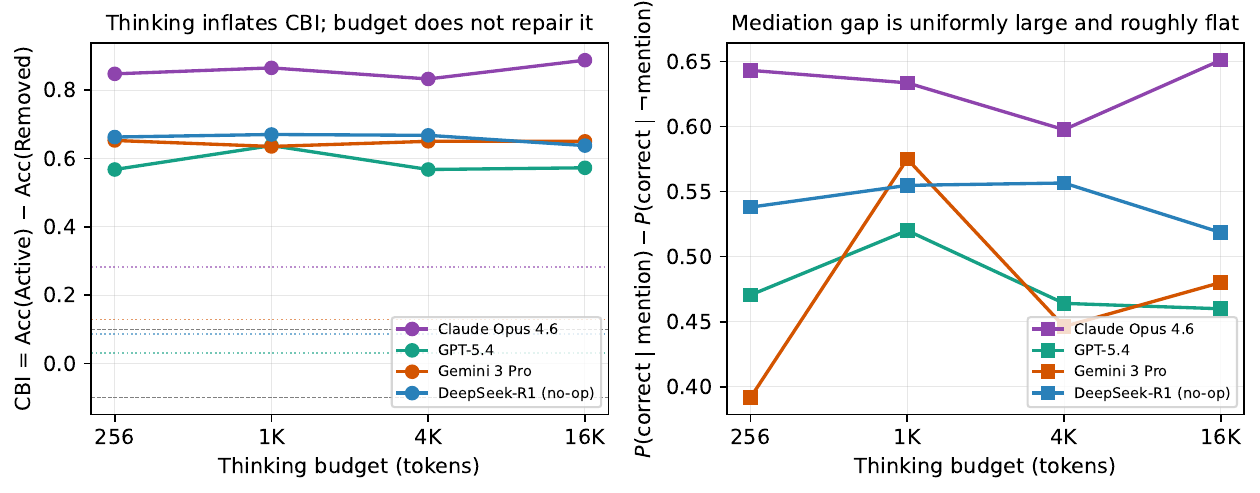}
\caption{Reasoning-budget sweep on four thinking-mode models.
\emph{Left:} CBI as a function of budget; the four dotted horizontals
show each model's zero-shot (non-thinking) baseline. Thinking-mode
moves all four models well above the zero-shot CBI; budget within
thinking-mode does not. \emph{Right:} the mediation conditional gap
$\Pr(\text{c}\mid\text{m})\!-\!\Pr(\text{c}\mid\neg\text{m})$ stays
uniformly large and budget-insensitive.}
\label{fig:budget}
\end{figure}

We sweep $\text{budget}\in\{256, 1024, 4096, 16{,}384\}$ thinking
tokens on the four thinking-mode models (Claude Opus~4.6, GPT-5.4,
Gemini~3~Pro, DeepSeek-R1) over the full \textsc{core100} quartet
($T\!=\!4$ trials per cell). Per-budget Active/Removed accuracy, CBI,
and mediation metrics are summarized in Fig.~\ref{fig:budget}.

Thinking-mode \emph{itself} moves all four models well into the
over-activation regime: at budget=$256$ tokens, CBI is $+0.85$ for
Claude Opus (vs.\ $+0.28$ zero-shot), $+0.57$ for GPT-5.4 (vs.\
$+0.03$), $+0.65$ for Gemini 3 Pro (vs.\ $+0.13$), $+0.66$ for
DeepSeek-R1 (vs.\ $+0.09$). Removed accuracy collapses ($\leq\!0.22$)
while Active accuracy holds within $0.05$ of zero-shot. Budget within
thinking-mode is essentially a no-op: CBI varies by $\leq 0.05$ across
the four-point grid. Mediation is uniformly strong:
$\Pr(\text{m})\!\in\![0.82,0.92]$, conditional gap
$\Pr(\text{c}\mid\text{m})-\Pr(\text{c}\mid\neg\text{m})\!\in\![+0.43,+0.67]$,
mediated-correctness share $\geq\!0.93$ throughout. Extra compute does
not change \emph{which} items get the mention, only the strength of
the bias. Per-budget detail is in Table~\ref{tab:budget_full}
(Appendix).

\subsection{Conditions K, S, R: open-weight probes}
\label{sec:results:K}

\paragraph{(K) Knowledge.} A linear probe on per-layer hidden states
decodes the binary Active-vs-Removed label well above chance for both
models (Table~\ref{tab:probe}). Qwen3-14B peaks at layer~27 with
$94.5\% \pm 1.9\%$ cross-validated accuracy; GPT-OSS-20B peaks at
layer~20 with $88.0\% \pm 4.3\%$. The constraint is internally
encoded in both models, well above the preregistered
$\theta_K\!=\!0.80$ threshold. We additionally observe broad-layer
accessibility: $\geq\!0.92$ accuracy across layers $25$--$36$ in
Qwen3-14B and $\geq\!0.85$ across layers $4$--$24$ in GPT-OSS-20B
(full per-layer curves in Table~\ref{tab:probe_layers_full},
Appendix).

\paragraph{(S) Symmetry.} Re-training the probe on a balanced
constraint-active/constraint-removed split shows the test-accuracy gap
between the two halves is within $\theta_S\!=\!0.05$ for both models
at their best layer ($\Delta\!=\!0.011$ for Qwen3-14B; $\Delta\!=\!0.024$
for GPT-OSS-20B). The knowledge is present symmetrically; an
asymmetry-based explanation of the behavioral CBI is ruled out.

\paragraph{(R) Routing.} The probe-projected magnitude at the best
layer correlates only weakly with the per-trial gold--shortcut logit
gap (Spearman $\rho\!=\!0.125$ for Qwen3-14B, $0.224$ for GPT-OSS-20B);
the per-layer maxima ($0.242$, $0.302$) also fall below the
preregistered $\theta_R\!=\!0.30$. The probe direction is present but
not strongly read by the decision head---the signature of a routing
failure.

\begin{table}[t]
\small
\centering
\begin{tabular}{@{}lcc@{}}
\toprule
& Qwen3-14B & GPT-OSS-20B \\
\midrule
Probe best layer & $L_{27}$ & $L_{20}$ \\
CV accuracy (K) & $0.945\pm 0.019$ & $0.880\pm 0.043$ \\
Sym.\ gap (S) & $0.011$ & $0.024$ \\
Routing $\rho$ @ best $L$ (R) & $0.125$ & $0.224$ \\
Max $\rho$ over layers (R) & $0.242$ & $0.302$ \\
\bottomrule
\end{tabular}
\caption{Conditions K, S, R on the two open-weight models. Both
satisfy K and S; both fall below $\theta_R=0.30$ at the probe's best
layer (Qwen3-14B clearly; GPT-OSS-20B marginally).}
\label{tab:probe}
\end{table}

\subsection{Condition P: activation patching}
\label{sec:results:P}

Patching the constraint-encoding hidden state from the Explicit prompt
into the Active prompt at the probe's best layer produces opposite
signatures in the two models (Table~\ref{tab:patch}, $n\!=\!100$
scenarios each).

\textbf{Qwen3-14B.} The Active gold--shortcut gap rises from $+1.40$
to $+7.78$~nats ($\Delta\!=\!+6.38$), an order-of-magnitude shift,
while Removed pairs are only mildly affected ($\Delta\!=\!-0.84$).
This is a clean repair signature: the constraint-encoding direction
\emph{exists} in the donor and \emph{causally} shifts the recipient's
decision when patched. \textbf{P is satisfied}.
\textbf{GPT-OSS-20B.} The same intervention yields no effect:
$\Delta\!=\!-0.069$ on Active patches, $+0.029$ on Removed (both within
the $\sim 3$-nat noise floor). Despite the $88\%$-accurate probe at
$L_{20}$, the constraint direction does not causally control the
decision. \textbf{P fails}.

\begin{figure}[t]
\centering
\includegraphics[width=\linewidth]{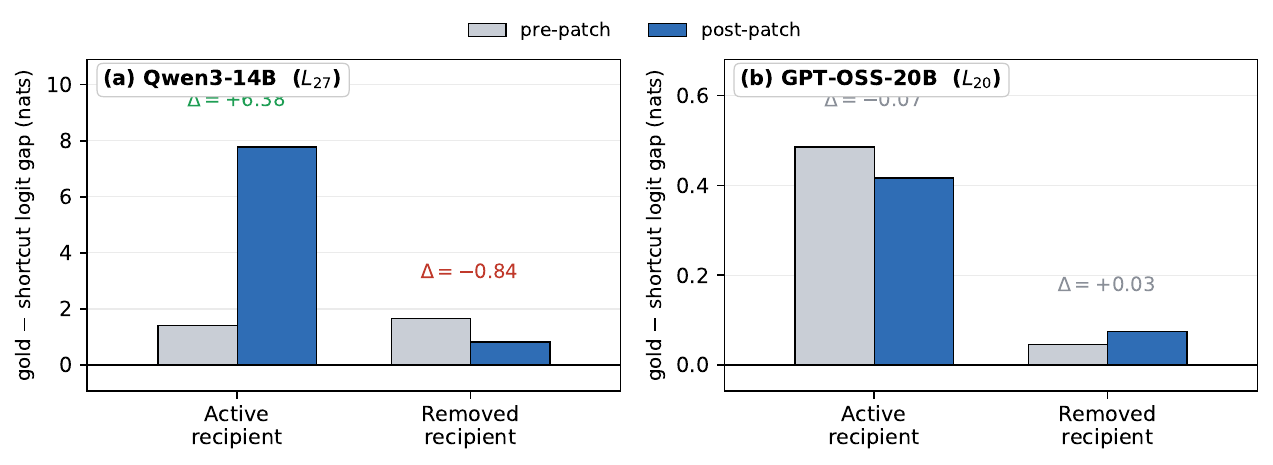}
\caption{Activation-patching deltas at the probe's best layer.
Pre-patch (grey) vs.\ post-patch (blue) mean gold--shortcut logit gap.
Qwen3-14B shows a clean repair signature on Active recipients
($\Delta=+6.38$) with controlled Removed deviation; GPT-OSS-20B shows
no effect, despite the strong probe at the same layer.}
\label{fig:patching}
\end{figure}

\begin{table}[t]
\small
\centering
\setlength{\tabcolsep}{4pt}
\begin{tabular}{@{}lrrr@{}}
\toprule
& pre-gap & patched & $\Delta$ (nats) \\
\midrule
\multicolumn{4}{@{}l}{\emph{Qwen3-14B, $L_{27}$ Explicit$\to$Recip.}} \\
Active recip.\  & $+1.398$ & $+7.781$ & $\mathbf{+6.382}$ \\
Removed recip.\ & $+1.660$ & $+0.820$ & $-0.840$ \\
\midrule
\multicolumn{4}{@{}l}{\emph{GPT-OSS-20B, $L_{20}$ Explicit$\to$Recip.}} \\
Active recip.\  & $+0.486$ & $+0.417$ & $\mathbf{-0.069}$ \\
Removed recip.\ & $+0.046$ & $+0.074$ & $+0.029$ \\
\bottomrule
\end{tabular}
\caption{Activation-patching repair test. Qwen3-14B exhibits a clean
repair signature on Active patches with controlled Removed harm.
GPT-OSS-20B exhibits no effect, despite a strong probe at the same
layer (K satisfied).}
\label{tab:patch}
\end{table}

\paragraph{K/S/R/P verdict.} Qwen3-14B is K\,\checkmark\,S\,\checkmark\,
R\,$\partial$\,P\,\checkmark (causally available, weakly routed);
GPT-OSS-20B is K\,\checkmark\,S\,\checkmark\,R\,$\partial$\,P\,$\times$
(textbook routing failure). The 7 over-activation models share this
K/S/R structure but add a strong constraint-respecting prior.

\section{Discussion}
\label{sec:discussion}

\subsection{Routing, not knowledge}
\label{sec:disc:routing}

The K/S/R/P measurements disconfirm the simplest account of
hidden-constraint failure---that the model does not know the
constraint. Probes decode constraint applicability at $94.5\%$
(Qwen3-14B, $L_{27}$) and $88.0\%$ (GPT-OSS-20B, $L_{20}$), broadly
across 10+ layers, and the symmetry test rules out a
constraint-active-only artifact. Knowledge \emph{is} there.

What separates the two open weights is the \emph{causal} test:
patching the donor Explicit activation repairs Qwen3-14B ($+6.4$ nats,
Removed unchanged) but does nothing in GPT-OSS-20B ($-0.07$)---the
direction is decodable yet unread by the decision head. The
implication: Active-item accuracy alone cannot tell competent inference
from a constraint-heavy prior; the Removed pair makes the distinction
visible.

\subsection{Why every prompted mitigation fails the same way}
\label{sec:disc:mitigations}

The mitigation-frontier evidence is the most surprising part of this
paper. CoT, precondition listing, goal decomposition, and even
counterfactual checking---an instruction to explicitly imagine the
salient resource absent---all produce the same harm profile: large
PairHarm ($+0.28$ to $+0.65$), small ActiveGain ($-0.08$ to $+0.11$),
$\geq 91\%$ mediated-correctness share. The strategies are not
differentiated by how they reason; they are differentiated only by
how often they trigger the \emph{same downstream behavior}---namely,
mentioning the hidden constraint.

A single common cause fits. Each strategy lifts $\Pr(\text{mention})$
from a low baseline ($0.3$--$0.5$) to a uniformly high value
($0.84$--$0.96$); whether or not the constraint applies, the model
talks about it, and the conservative answer then dominates. The
conditional gap is large ($+0.34$ to $+0.64$) but \emph{symmetric}: on
Removed items, mention flips the answer the wrong way, netting to
near-zero on Pair-Consistent Accuracy---yet the literature reports only
the constraint-active half. Prompted reasoning cannot \emph{decide}
when to mention the constraint, only globally bias the rate; selective
activation is a routing problem, not a prompt-level property.

\subsection{Two failure modes, two remedies}
\label{sec:disc:two_modes}

The mechanistic measurements suggest different remedies per population.
\textbf{Over-activation} (7 models) combines intact routing with a
prior bias (large SalAdjGain); prompted mitigations push exactly the
coefficient that already over-fires. \textbf{Balanced} (5 models) is
where Qwen3-14B's patching success shows a \emph{selective}
intervention---inject the constraint direction only when it applies---
is mechanistically feasible. \textbf{Under-activation} (2 models) is
the hard case: GPT-OSS-20B's direction exists at $L_{20}$ but is unread
by the decision head, and the single-layer patch is insufficient. A
\emph{learned activation gate} that decides per item when to patch
would test whether circuit-level intervention can reach the repair
corner; the released probe weights, cross-architecture replication, and
a multi-label extension to multi-constraint scenarios are the natural
next steps.

\section{Conclusion}
\label{sec:conclusion}

We recast hidden-constraint failure in LLMs as \emph{conditional
activation}: the constraint is decodable from hidden state (K)
symmetrically (S) but not reliably routed into the decision (R).
Patching dissociates the two failure modes, and a mitigation frontier
shows every prompted intervention inflates conservative bias rather
than repairing routing---so the fix must target the routing direction
itself.

\section*{Limitations}
\label{sec:limitations}

\paragraph{Scenario count.} \textsc{core100} stratifies across the
HOB cell taxonomy at the cost of statistical power per cell. Our
between-condition contrasts use the matched paired design to recover
power, but per-cell error bars remain wide for finer-grained slicing.

\paragraph{Two open weights for the causal claim.} The
K/S/R/P mechanism is established on Qwen3-14B and GPT-OSS-20B only.
While the behavioral signature extends to all 14 models, the causal
patching evidence is backed by two open-weight models, chosen for
release-license reasons and to span the two failure modes.

\paragraph{Single-layer patching.} Our Repair condition is satisfied
by patching at the probe's single best layer. A multi-layer or
block-wise intervention may reveal repair in models where the
single-layer patch fails (e.g.\ GPT-OSS-20B), and our analysis
therefore underestimates the rate at which Repair holds.

\paragraph{English, single-domain.} HOB scenarios are English
everyday-domain. Cross-lingual replication and domain-shifted versions
(e.g.\ medical, legal) are open questions; we expect the K/S/R/P
formalism to transfer but the failure-mode partition may not.

\paragraph{Prompted-mitigation coverage.} We evaluate four prompted
strategies that span the published taxonomy of pragmatic-reasoning
interventions. Reinforcement-learned
mitigations~\citep{lambert2024rlhf,gulcehre2023reinforced} and
fine-tuned debiasing remain untested. Our prediction, based on the
mediation chain, is that any intervention that operates on the trace
surface will fall in the same harm-without-repair region; we leave
that to empirical verification.

\paragraph{Judge model.} All 54{,}400 quartet trials and 128{,}000
mitigation trials are judged by a single Qwen3-32B local model with
deterministic decoding. This is a standard choice in the
heuristic-shortcut literature, but a single-judge evaluation may
systematically miss certain failure modes (e.g.\ correct answers given
for the wrong reason). We release the judging prompt and all raw
generations so reanalysis with alternative judges is straightforward.

\paragraph{API model versioning.} Frontier models on closed APIs
(Claude, GPT, Gemini, Grok, Kimi, DeepSeek) may be updated between
data collection (May 2026) and publication. We pin exact model
identifiers in Table~\ref{tab:model_versions}
(Appendix~\ref{app:model_versions}) and release the full raw
generations, but pin-violation risk on closed APIs is non-zero.

\paragraph{Heuristic prerequisite-mention detection.} Our mediation
analysis uses content-word overlap with the canonical hidden
constraint to detect ``the model mentions the prerequisite.'' This
catches cases where the model paraphrases the constraint but may miss
implicit references; we report mediation conservatively.

\bibliography{anthology}

\clearpage
\onecolumn
\appendix

\section{Model versions}
\label{app:model_versions}

Table~\ref{tab:model_versions} pins the exact model identifiers and
access modality for all 14 evaluated models. All API data were
collected in May 2026; closed-API models may be updated by their
providers after this date. Open-weight models are served locally from
the listed Hugging Face checkpoints and are version-stable.

\begin{table}[H]
\footnotesize
\centering
\setlength{\tabcolsep}{4pt}
\begin{tabular}{@{}l l p{3.4cm}@{}}
\toprule
\textbf{Model} & \textbf{Provider} & \textbf{Identifier} \\
\midrule
GPT-5.2          & OpenAI   & \texttt{gpt-5.2} \\
GPT-5.4          & OpenAI   & \texttt{gpt-5.4} \\
Claude Opus 4.6  & Anthropic& \texttt{claude-opus-4-6} \\
Claude Sonnet 4.5& Anthropic& \texttt{claude-sonnet-4-5-\allowbreak 20250929} \\
DeepSeek-R1      & DeepSeek & \texttt{deepseek-reasoner} \\
Gemini 3 Pro     & Google   & \texttt{gemini-3.1-pro-\allowbreak preview} \\
Grok 4.2         & xAI      & \texttt{grok-4.20-beta-\allowbreak 0309-reasoning} \\
Kimi K2.5        & Moonshot & \texttt{kimi-k2.5} \\
Llama-4 Scout    & Groq     & \texttt{meta-llama/\allowbreak llama-4-scout-\allowbreak 17b-16e-instruct} \\
GPT-OSS-120B     & Groq     & \texttt{openai/\allowbreak gpt-oss-120b} \\
\midrule
Qwen3-14B        & local    & \texttt{Qwen/Qwen3-14B} \\
Qwen3-32B        & local    & \texttt{Qwen/Qwen3-32B} \\
Qwen3.5-27B      & local    & \texttt{Qwen/Qwen3.5-27B} \\
GPT-OSS-20B      & local    & \texttt{openai/\allowbreak gpt-oss-20b} \\
\bottomrule
\end{tabular}
\caption{Exact model identifiers and access modality (API vs.\ local
Hugging Face checkpoint). API data collected May 2026. The judge model
is Qwen3-32B for all trials. The four local checkpoints (bottom block)
are also the open weights used for hidden-state probing and activation
patching (Qwen3-14B, GPT-OSS-20B).}
\label{tab:model_versions}
\end{table}

\section{Per-(model, strategy) mitigation frontier}
\label{app:mit_frontier_full}

Table~\ref{tab:mit_frontier_full} reports the full $10\times 4 = 40$
mitigation cells (10 API models, 4 prompted strategies) referenced in
\S\ref{sec:results:frontier}. All values are relative to each model's
zero-shot baseline on the same 100 \textsc{core100} scenarios.

\begin{table}[H]
\small
\centering
\setlength{\tabcolsep}{3.5pt}
\begin{tabular}{llrrrrrr}
\toprule
\textbf{Strategy} & \textbf{Model} &
  \textsc{ActGain} & \textsc{PairHarm} & $\Delta$CBI &
  $\Pr(\text{c}\mid\text{m})$ & $\Pr(\text{c}\mid\neg\text{m})$ & $\Pr(\text{m})$ \\
\midrule
\multicolumn{8}{l}{\emph{cot}} \\
cot & claude-opus-4.6   & $+0.033$ & $+0.404$ & $+0.437$ & 0.759 & 0.182 & 0.928 \\
cot & claude-sonnet-4.5 & $+0.073$ & $+0.450$ & $+0.523$ & 0.764 & 0.124 & 0.935 \\
cot & deepseek-r1       & $+0.035$ & $+0.569$ & $+0.604$ & 0.786 & 0.270 & 0.848 \\
cot & gemini-3-pro      & $+0.003$ & $+0.504$ & $+0.507$ & 0.766 & 0.312 & 0.957 \\
cot & gpt-5.2           & $-0.023$ & $+0.339$ & $+0.316$ & 0.751 & 0.409 & 0.852 \\
cot & gpt-5.4           & $+0.004$ & $+0.468$ & $+0.473$ & 0.767 & 0.348 & 0.849 \\
cot & gpt-oss-120b      & $+0.050$ & $+0.637$ & $+0.687$ & 0.683 & 0.079 & 0.924 \\
cot & grok-4.2          & $-0.003$ & $+0.607$ & $+0.604$ & 0.792 & 0.365 & 0.839 \\
cot & kimi-k2.5         & $+0.001$ & $+0.424$ & $+0.424$ & 0.792 & 0.157 & 0.942 \\
cot & llama-4           & $-0.057$ & $+0.353$ & $+0.295$ & 0.642 & 0.124 & 0.927 \\
\midrule
\multicolumn{8}{l}{\emph{precondition\_listing}} \\
prec & claude-opus-4.6  & $+0.056$ & $+0.351$ & $+0.408$ & 0.788 & 0.241 & 0.946 \\
prec & claude-sonnet-4.5 & $+0.089$ & $+0.399$ & $+0.489$ & 0.775 & 0.292 & 0.944 \\
prec & deepseek-r1      & $+0.061$ & $+0.534$ & $+0.595$ & 0.775 & 0.259 & 0.934 \\
prec & gemini-3-pro     & $+0.004$ & $+0.404$ & $+0.408$ & 0.799 & 0.447 & 0.959 \\
prec & gpt-5.2          & $+0.022$ & $+0.279$ & $+0.301$ & 0.771 & 0.424 & 0.917 \\
prec & gpt-5.4          & $+0.084$ & $+0.447$ & $+0.531$ & 0.798 & 0.336 & 0.911 \\
prec & gpt-oss-120b     & $+0.064$ & $+0.544$ & $+0.608$ & 0.706 & 0.191 & 0.944 \\
prec & grok-4.2         & $+0.057$ & $+0.566$ & $+0.623$ & 0.799 & 0.306 & 0.927 \\
prec & kimi-k2.5        & $+0.009$ & $+0.367$ & $+0.377$ & 0.814 & 0.333 & 0.954 \\
prec & llama-4          & $-0.018$ & $+0.336$ & $+0.318$ & 0.691 & 0.178 & 0.900 \\
\midrule
\multicolumn{8}{l}{\emph{goal\_decomposition}} \\
goal & claude-opus-4.6  & $+0.016$ & $+0.354$ & $+0.370$ & 0.761 & 0.261 & 0.941 \\
goal & claude-sonnet-4.5 & $+0.099$ & $+0.397$ & $+0.496$ & 0.784 & 0.294 & 0.942 \\
goal & deepseek-r1      & $+0.039$ & $+0.565$ & $+0.604$ & 0.757 & 0.264 & 0.912 \\
goal & gemini-3-pro     & $+0.030$ & $+0.493$ & $+0.523$ & 0.784 & 0.353 & 0.964 \\
goal & gpt-5.2          & $+0.021$ & $+0.325$ & $+0.346$ & 0.760 & 0.386 & 0.903 \\
goal & gpt-5.4          & $+0.059$ & $+0.481$ & $+0.540$ & 0.779 & 0.294 & 0.913 \\
goal & gpt-oss-120b     & $+0.076$ & $+0.596$ & $+0.672$ & 0.688 & 0.120 & 0.951 \\
goal & grok-4.2         & $+0.041$ & $+0.609$ & $+0.649$ & 0.794 & 0.260 & 0.913 \\
goal & kimi-k2.5        & $+0.017$ & $+0.406$ & $+0.423$ & 0.809 & 0.248 & 0.948 \\
goal & llama-4          & $-0.061$ & $+0.359$ & $+0.298$ & 0.644 & 0.123 & 0.927 \\
\midrule
\multicolumn{8}{l}{\emph{counterfactual\_check (oracle)}} \\
cf  & claude-opus-4.6   & $+0.023$ & $+0.323$ & $+0.345$ & 0.787 & 0.300 & 0.916 \\
cf  & claude-sonnet-4.5 & $+0.077$ & $+0.371$ & $+0.448$ & 0.797 & 0.205 & 0.908 \\
cf  & deepseek-r1       & $+0.031$ & $+0.513$ & $+0.544$ & 0.801 & 0.296 & 0.854 \\
cf  & gemini-3-pro      & $+0.019$ & $+0.425$ & $+0.445$ & 0.810 & 0.345 & 0.936 \\
cf  & gpt-5.2           & $-0.008$ & $+0.324$ & $+0.316$ & 0.777 & 0.376 & 0.865 \\
cf  & gpt-5.4           & $+0.006$ & $+0.472$ & $+0.478$ & 0.772 & 0.380 & 0.845 \\
cf  & gpt-oss-120b      & $+0.109$ & $+0.548$ & $+0.657$ & 0.730 & 0.177 & 0.940 \\
cf  & grok-4.2          & $+0.029$ & $+0.645$ & $+0.674$ & 0.796 & 0.286 & 0.863 \\
cf  & kimi-k2.5         & $-0.003$ & $+0.405$ & $+0.402$ & 0.815 & 0.302 & 0.912 \\
cf  & llama-4           & $-0.082$ & $+0.339$ & $+0.257$ & 0.664 & 0.212 & 0.845 \\
\bottomrule
\end{tabular}
\caption{Per-(strategy, model) mitigation frontier with mediation
detail. \textsc{ActGain} is the change in Active accuracy relative to
zero-shot; \textsc{PairHarm} is the change in Removed accuracy
expressed as CBI inflation; $\Delta$CBI is \textsc{ActGain}\,+\,
\textsc{PairHarm}; $\Pr(\text{c}\mid\text{m})$ and
$\Pr(\text{c}\mid\neg\text{m})$ are correctness conditional on
prerequisite mention; $\Pr(\text{m})$ is the marginal mention rate.}
\label{tab:mit_frontier_full}
\end{table}

\section{Per-budget breakdown}
\label{app:budget_full}

Table~\ref{tab:budget_full} reports per-(model, budget) Active accuracy,
Removed accuracy, CBI, mention rate, and mediated correctness share for
the four thinking-mode models referenced in \S\ref{sec:results:budget}.

\begin{table}[H]
\footnotesize
\centering
\setlength{\tabcolsep}{3.5pt}
\begin{tabular}{@{}lrrrrrr@{}}
\toprule
Model & B & A & R & CBI & $\Pr(\text{m})$ & MedSh \\
\midrule
claude-o & 256 & 0.89 & 0.04 & $+0.85$ & 0.87 & 0.99 \\
claude-o & 1K  & 0.87 & 0.01 & $+0.87$ & 0.87 & 0.99 \\
claude-o & 4K  & 0.87 & 0.03 & $+0.83$ & 0.87 & 0.97 \\
claude-o & 16K & 0.87 & -0.02 & $+0.89$ & 0.88 & 0.99 \\
deepseek & 256 & 0.83 & 0.17 & $+0.66$ & 0.85 & 0.94 \\
deepseek & 1K  & 0.86 & 0.19 & $+0.67$ & 0.85 & 0.94 \\
deepseek & 4K  & 0.84 & 0.17 & $+0.67$ & 0.87 & 0.95 \\
deepseek & 16K & 0.83 & 0.20 & $+0.64$ & 0.87 & 0.93 \\
gemini-3 & 256 & 0.87 & 0.22 & $+0.65$ & 0.90 & 0.95 \\
gemini-3 & 1K  & 0.83 & 0.20 & $+0.64$ & 0.92 & 0.97 \\
gemini-3 & 4K  & 0.84 & 0.19 & $+0.65$ & 0.91 & 0.95 \\
gemini-3 & 16K & 0.87 & 0.21 & $+0.65$ & 0.92 & 0.96 \\
gpt-5.4 & 256 & 0.78 & 0.21 & $+0.57$ & 0.84 & 0.93 \\
gpt-5.4 & 1K  & 0.81 & 0.17 & $+0.64$ & 0.83 & 0.95 \\
gpt-5.4 & 4K  & 0.79 & 0.22 & $+0.57$ & 0.83 & 0.93 \\
gpt-5.4 & 16K & 0.79 & 0.22 & $+0.57$ & 0.82 & 0.93 \\
\bottomrule
\end{tabular}
\caption{Per-budget breakdown of the reasoning-budget sweep. B is in
tokens; CBI is signed ($>\!+0.10$ = over-activation).}
\label{tab:budget_full}
\end{table}

\section{Per-layer probe accuracy}
\label{app:probe_layers_full}

Table~\ref{tab:probe_layers_full} reports cross-validated probe
accuracy at every transformer layer for both open-weight models. Bold
marks the best layer used for downstream routing correlation
(\S\ref{sec:results:K}) and activation patching (\S\ref{sec:results:P}).

\begin{table}[H]
\footnotesize
\centering
\setlength{\tabcolsep}{3.5pt}
\begin{tabular}{@{}rcc|rcc@{}}
\toprule
L & Qwen & GPT-OSS & L & Qwen & GPT-OSS \\
 & 14B & 20B & & 14B & 20B \\
\midrule
0 & 0.595 & 0.560 & 21 & 0.930 & \textbf{0.875} \\
1 & 0.715 & 0.730 & 22 & 0.930 & 0.870 \\
2 & 0.770 & 0.820 & 23 & 0.935 & 0.870 \\
3 & 0.800 & 0.835 & 24 & 0.940 & 0.860 \\
4 & 0.815 & 0.875 & 25 & 0.940 & -- \\
5 & 0.820 & 0.860 & 26 & 0.940 & -- \\
6 & 0.830 & 0.860 & 27 & \textbf{0.945} & -- \\
7 & 0.835 & 0.880 & 28 & 0.940 & -- \\
8 & 0.835 & 0.860 & 29 & 0.935 & -- \\
9 & 0.840 & 0.860 & 30 & 0.935 & -- \\
10 & 0.840 & 0.860 & 31 & 0.930 & -- \\
11 & 0.850 & 0.870 & 32 & 0.930 & -- \\
12 & 0.855 & 0.870 & 33 & 0.935 & -- \\
13 & 0.870 & 0.875 & 34 & 0.930 & -- \\
14 & 0.870 & 0.875 & 35 & 0.930 & -- \\
15 & 0.880 & 0.870 & 36 & 0.930 & -- \\
16 & 0.900 & 0.870 & 37 & 0.920 & -- \\
17 & 0.910 & 0.860 & 38 & 0.920 & -- \\
18 & 0.920 & 0.860 & 39 & 0.905 & -- \\
19 & 0.925 & 0.870 & 40 & 0.905 & -- \\
20 & 0.930 & \textbf{0.880} &    &       &       \\
\bottomrule
\end{tabular}
\caption{Per-layer cross-validated probe accuracy on the Active vs.\
Removed binary label, scenario-grouped 5-fold. Standard deviations
range from 0.018 to 0.052 (omitted for compactness). Bold = layer
chosen for downstream patching. Qwen3-14B has 41 layers, GPT-OSS-20B
has 25; the right column for layers $\geq 25$ is empty for
GPT-OSS-20B.}
\label{tab:probe_layers_full}
\end{table}

\section{Hint-ladder details}
\label{app:ladder}

\begin{figure}[H]
\centering
\includegraphics[width=\linewidth]{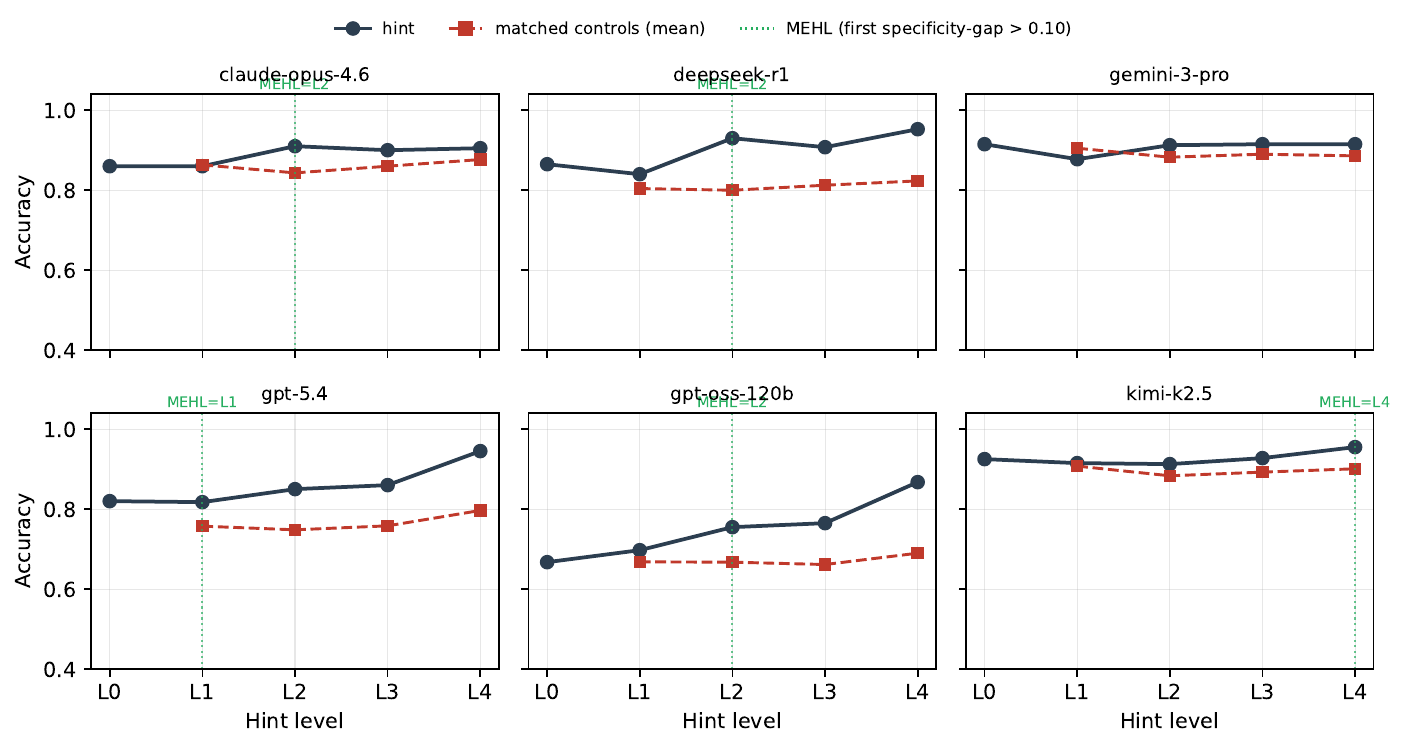}
\caption{Hint-ladder activation curves with matched negative controls
on six API models. Hint accuracy (solid black) climbs monotonically;
the mean of the three matched controls (dashed red) stays flat,
demonstrating the gain is hint-specific rather than salience-driven.}
\label{fig:ladder}
\end{figure}

\begin{figure}[H]
\centering
\includegraphics[width=\linewidth]{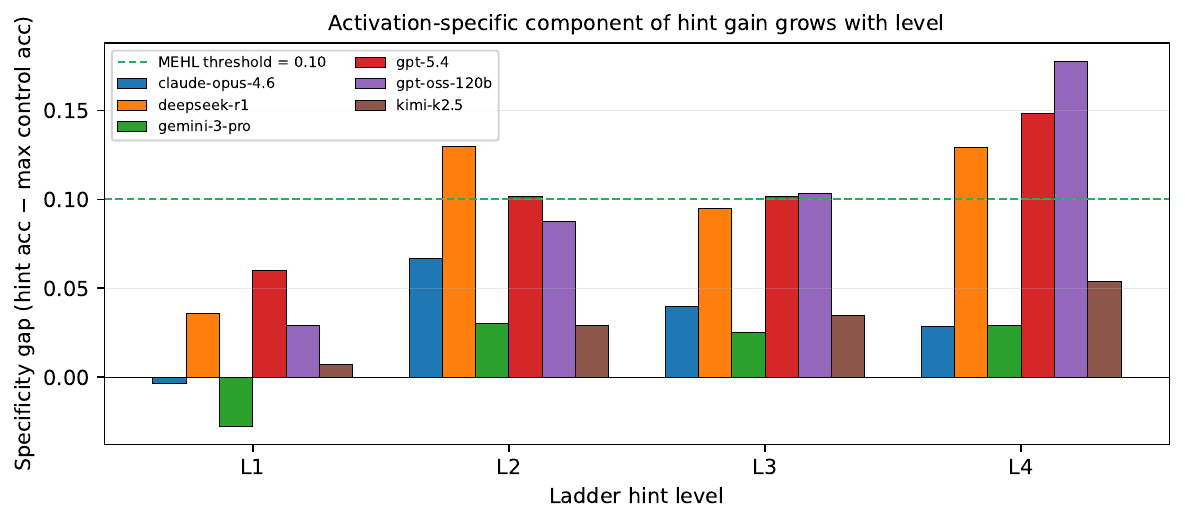}
\caption{Hint-ladder specificity gap per level (hint accuracy minus
best matched control). Reaches MEHL threshold (green dashed) at $L_2$
or $L_3$ for most models.}
\label{fig:ladder_specgap}
\end{figure}

Fig.~\ref{fig:ladder} reports the per-model hint-ladder activation
curve with matched negative controls; Fig.~\ref{fig:ladder_specgap}
shows the per-level specificity gap. The activation-specific component
of the hint gain grows monotonically with hint level on every
ladder-complete model.

\section{Per-layer probe accuracy curve}
\label{app:probe_layers_fig}

\begin{figure}[H]
\centering
\includegraphics[width=\linewidth]{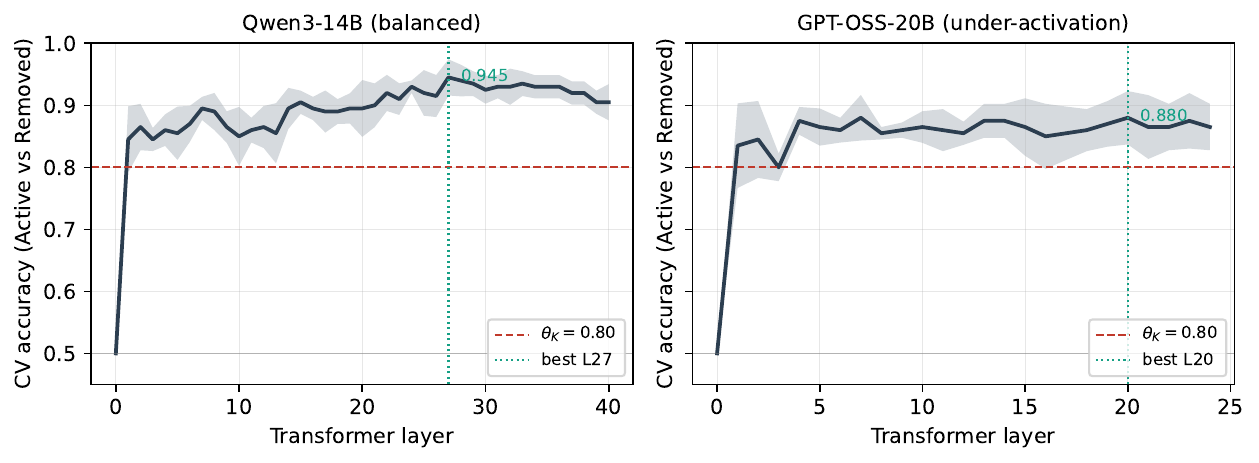}
\caption{Per-layer linear-probe CV accuracy on the Active vs.\ Removed
label. Shaded band is $\pm 1\,\mathrm{SD}$ over scenario-grouped folds;
red dashed at $\theta_K=0.80$; green dotted marks best layer.}
\label{fig:probe_layers}
\end{figure}

Fig.~\ref{fig:probe_layers} visualizes the per-layer probe accuracy
reported numerically in Table~\ref{tab:probe_layers_full}.

\section{Methods detail: reproducibility}
\label{app:repro}

\paragraph{Code and data release.} All prompts, the salience-control
bank, mitigation prefixes, ladder templates, probe and patching code,
and metric implementations will be released. We seed ladder
construction (deterministic hash by \texttt{instance\_id}) and probe
training (\texttt{numpy} seed $42$); all stochasticity is at the
model-API level. The full quartet, ladder, mitigation, and budget
collection is resume-safe by $(\text{instance\_id},$ $\text{trial},$
$\text{strategy},$ $\text{budget})$ keys, with retry passes that
preserve the original prompt prefix and thinking budget so re-runs
are exchangeable with first-time runs.

\paragraph{Provider-specific budget routing.} Anthropic models accept
\texttt{thinking}=\{\texttt{enabled}, \texttt{budget\_tokens}=$b$\}
with $b$ clamped to a minimum of $1{,}024$ tokens. OpenAI o-series and
GPT-5 series accept \texttt{reasoning\_effort}$\in$\{\texttt{low},
\texttt{medium}, \texttt{high}\}, mapped from $b$ via thresholds
$1024$ and $4096$. Gemini accepts
\texttt{ThinkingConfig(thinking\_budget}=$b$\texttt{)}. DeepSeek-R1
emits chain-of-thought traces but does not expose a budget knob; the
requested $b$ is recorded for parity but is a no-op.

\paragraph{Activation patching mechanics.} For each scenario, we
forward the donor (\textsc{Explicit}) prompt and cache the
final-token hidden state at the probe's best layer $\ell^\star$ via
a PyTorch forward hook on \texttt{model.model.layers[$\ell^\star$]}.
We then forward the recipient prompt, with a second hook that
overwrites the layer-output tensor at the recipient's last position
with the cached donor activation; logits are extracted at the same
position and the gold--shortcut log-probability gap is recorded. The
gap difference $\Delta = \Delta^{\text{patched}} -
\Delta^{\text{pre}}$ is reported as the patching effect.

\paragraph{Prerequisite-mention detection.} A model response is
classified as ``mentioning the prerequisite'' iff it shares
$\geq\!2$ distinct content-word matches with the canonical hidden
constraint string, after lowercasing, stripping punctuation, and
filtering a small stoplist of high-frequency English function words.
This is a conservative, lexical-overlap definition; it may
undercount paraphrastic mentions, which would only strengthen the
mediated-correctness-share lower bounds reported in
\S\ref{sec:results:frontier}.

\paragraph{Judge.} All $54{,}400 + 128{,}000 + 25{,}600 = 208{,}000$
trials are judged by Qwen3-32B with deterministic decoding, using the
multiple-choice extraction prompt from \citet{hob2026}. Trials with
API errors at collection are retried with exponential backoff;
verdicts for retried trials are recomputed by the same judge to
maintain consistency.

\end{document}